\documentclass[conference]{IEEEtran}

\usepackage{times}
\usepackage[numbers]{natbib}
\usepackage{booktabs} 
\usepackage{graphicx}
\usepackage{tabularx}
\usepackage{caption}
\usepackage{subcaption}
\usepackage{algorithm}
\usepackage{algpseudocode}
\usepackage{subscript}
\usepackage{color}
\usepackage{wrapfig}
\usepackage[bookmarks=true]{hyperref}
\usepackage{verbatim}
\usepackage{multicol}

\graphicspath{ {fig/} }

\newcommand{\savefootnote}[2]{\footnote{\label{#1}#2}}

\newcolumntype{L}[1]{>{\raggedright\arraybackslash}p{#1}}
\newcolumntype{C}[1]{>{\centering\arraybackslash}p{#1}}
\newcolumntype{R}[1]{>{\raggedleft\arraybackslash}p{#1}}

\newcounter{myenum}
 {\end{list}}

\newenvironment{flushitemize}{%
\begin{list}{$\bullet$}
   {\setlength{\leftmargin}{15pt}}%
    \setlength{\labelwidth}{20pt}
    \setlength{\itemindent}{7pt}
    \setlength{\labelsep}{0.5em}
 \setlength{\itemsep}{1pt}
 \setlength{\parskip}{0pt}
 \setlength{\parsep}{0pt}}
 {\end{list}}

\begin{document}
\title{Effects of Interruptibility-Aware Robot Behavior}

\author{
    \authorblockN{Siddhartha Banerjee\IEEEauthorrefmark{1}, Andrew Silva\IEEEauthorrefmark{1}, Karen Feigh, Sonia Chernova}
    \authorblockA{\{siddhartha.banerjee, andrew.silva, karen.feigh, chernova\}@gatech.edu}
    \authorrefmark{1}Authors contributed equally to this work
}

\maketitle

\begin{abstract}
As robots become increasingly prevalent in human environments, there will inevitably be times when a robot needs to interrupt a human to initiate an interaction.  Our work introduces the first interruptibility-aware mobile robot system, and evaluates the effects of interruptibility-awareness on human task performance, robot task performance, and on human interpretation of the robot's social aptitude. Our results show that our robot is effective at predicting interruptibility at high accuracy, allowing it to interrupt at more appropriate times.  Results of a large-scale user study show that while participants are able to maintain task performance even in the presence of interruptions, interruptibility-awareness improves the robot's task performance and improves participant social perception of the robot. 

\end{abstract}

\IEEEpeerreviewmaketitle

\section{Introduction}
\label{sec:introduction}

Interruptions are distracting, potentially leading to task performance penalties~\cite{speier1997effects,McFarlane2002a}, stress~\cite{McFarlane2002a,Mark2008}, antipathy~\cite{Mutlu2008}, and even catastrophe~\cite{Sarter2013,Rivera2014}, depending on context. Therefore, a large body of work in human factors engineering (HFE) and human-computer interaction (HCI) research has studied interruptions and ways of mitigating their effects. Prior research has specifically identified the appropriateness of the \textit{timing of an interruption} as one of the most important factors dictating its consequences~\cite{miyata1986psychological,speier1997effects,McFarlane2002a,Rivera2014}. The appropriateness of timing has been referred to as \textit{interruptibility}~\cite{stern2011preliminary} and it is itself the focus of much research~\cite{Turner2015}. Low interruptibility signifies a person's desire to not be disturbed, while high interruptibility signifies that the person could be amenable to an interruption.

Today's robots have no interruptibility awareness, despite the fact that interactive robots are increasingly becoming deployed in human environments.  Many robot control architectures being developed in the research community for interactive applications enable robots to not only follow human instructions, but also to actively engage with a person to offer a service
\cite{brvsvcic2017you} or to ask for help \cite{fischer2014initiating,Rosenthal2012}.  As a result, robots performing deliveries, taking store inventory, organizing warehouses, and collaboratively working alongside humans on factory production lines increasingly have the potential to interrupt people, without any measure of the appropriateness or costs of such interruptions.  Extrapolating results from HCI research~\cite{McFarlane2002} to the domain of embodied robot interactions, suggests that inappropriate interruptions may have significant effects on many factors, including
\begin{flushitemize}
\item negatively impacting human task performance, if people are interrupted at inappropriate times,
\item negatively impacting robot task performance, as the robot wastes time attempting to interact with a person not receptive to the interaction, and
\item negatively impacting a person's social perception of the robot, and ultimately their willingness to use it.
\end{flushitemize}

\begin{figure}[t]
\centering
\includegraphics[width=\linewidth]{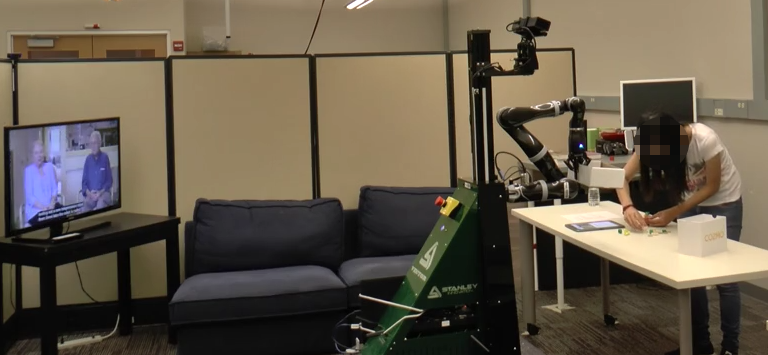}
\caption{The robot interrupts a participant in a building task.}
\label{fig:study-moment}
\vspace{-0.5cm}
\end{figure}

In order to develop robots that appropriately handle interruptions, it is important to determine \textit{when} a robot should interrupt, and \textit{how} it should behave during an interruption.  Prior work has explored how a robot should behave during interruptions by studying multiple approaches for engaging people~\cite{Saulnier2011,Chiang2014a}.  In this paper, we address the former question, introducing a self-contained interruptibility-aware mobile robot system and presenting a detailed analysis of the effects of interruptibility-aware behavior on the factors listed above.



Specifically, our work makes three contributions. First, we present the first integrated robotic system for autonomous classification of human interruptibility; our work uses multiple state-of-the-art computer vision detectors to extract contextual and social cues from visual data, with no instrumentation of the user or the environment.  Second, we evaluate the impact that interruptibility-aware robot behavior has on the task performance of interrupted humans and of the robot itself. Third, we evaluate the effect that interruptibility-aware robot behavior has on the humans' perceived suitability of the robot's social behavior. The second and third contributions are derived from a user study conducted with 42 participants in a mock manufacturing environment.

Our results show that (1) our integrated system is effective at predicting interruptibility at high accuracy, (2) an interruptibility-aware robot interrupts less often but at more appropriate times thereby increasing its efficiency, (3) confirming results from HFE~\cite{Kolbeinsson2017}, there is no significant benefit to human task throughput as a result of better interruption timing, and (4) users rate an interruptibility-aware robot as being more considerate. These results highlight key findings for face-to-face robot interruptions, underscore the social and task benefits of interruptibility-aware robot behaviours, and present directions for future research.

\section{Related Work}
\label{sec:related-work}


Existing work in robotics has modeled interruptibility in one of two ways. The first category of techniques relies on \textit{task} or \textit{contextual} knowledge.  For example cognitive architectures such as ACT-R/E~\cite{Trafton2013,Trafton2012} have been used to predict if humans might need assistance in resuming a task post-interruption by another human, a technique that easily extends to determining a moment to interrupt.  However, the above approach requires domain knowledge about a human task and constant surveillance of its execution, which is often unavailable to a general-purpose mobile robot deployment.  Others have modeled human availability based on past room occupancy, assuming that an open office door indicates the occupant's willingness to be interrupted~\cite{Rosenthal2012}, but this assumption ignores both social cues and task state to greatly simplify the interruptibility problem.

The second category of techniques relies on immediate \textit{social} cues that are independent of knowing the task or state of the human, thereby resulting in methods more easily applicable across a wider set of applications.  Most existing work in this category has relied on external sensors, such as motion capture systems, ground-mounted LIDAR, and ceiling cameras~\cite{Satake2009,Kato2015,Shi2015,brvsvcic2017you}.  Such systems can however be expensive and difficult to deploy in support of mobile robots traversing a large space.  Our work therefore focuses on using only onboard robot sensors.  Nigam \& Riek~\cite{Nigam2015} use onboard sensors for low-level audio-visual descriptors---such as GIST~\cite{oliva2001modeling} features and audio frequency \& volume features---as cues to context in classifying interruptibility (termed an \textit{appropriateness function}) on their collected dataset. We leverage advances in computer vision to gather high-level information from a scene, building on our prior work in interruptibility classification with onboard sensors~\cite{banerjee2017temporal}. In our work, we draw on social cues used in social engagement research~\cite{Bohus2009,Mollaret2016} and context cues in HCI interruptibility research~\cite{Fogarty2005,Sykes2014} to compare the interruptibility classification performance of various temporal models on a fixed image dataset.  We leverage the best computational model from \cite{banerjee2017temporal}, describe how we improve its performance on real-world human-robot interaction, and contribute an analysis of the effects interruptibility-aware robot behavior has on human task performance, robot task performance, and social perception of the robot.

The evaluation of face-to-face robot interruptions, in which a robot is co-present with the human, has been limited to qualitative measures to gauge the effectiveness of the interruption.  In~\cite{Saulnier2011}, the authors base their evaluation on participant self-assessed ``interruptedness'', while in~\cite{Chiang2014a}, the authors evaluate whether an interruption successfully captured the attention of a participant, without consideration for the appropriateness of interruption timing.  There is no prior work that quantitatively studies the \textit{task effects} of embodied robot interruptions on human and robot performance.  Interestingly, research from HCI and HFE disagree on the potential costs of interruptions.  HCI research has shown that people subject to on-screen mistimed interruptions experience significant loss in task performance~\cite{McFarlane2002,Adamczyk2004}, but recent results from HFE show that such loss is not noticeable with tasks that are embodied or skill-based, even when the interruptions might be computer mediated as in \cite{Lee2015,Kolbeinsson2017}. \cite{Kolbeinsson2017}, in particular, reason that performance loss is absent in an embodied setting because it is impossible to occlude the main task, which allows people to optimize common sub tasks and choose when to switch to an interruption.  In this work, we explore whether the results from HFE research generalize to robotic systems.


\section{Overview}
\label{sec:overview}

Our research seeks to develop interruptibility-awareness in robots and to evaluate the effects of this capability on human task performance, robot task performance, and on the human's interpretation of the robot's social aptitude. Specifically, we focus on the following research questions:

\begin{flushitemize}
\item[\textbf{RQ1}] Can an integrated system be developed to accurately estimate human interruptibility online on a robot platform?

\item[\textbf{RQ2}] How does interruptibility-aware robot behavior affect human task performance when a robot regularly needs assistance?

\item[\textbf{RQ3}] How does interruptibility-aware robot behavior affect robot task performance when relying on humans for assistance?

\item[\textbf{RQ4}] Does a robot appear more socially adept if it interrupts humans at appropriate moments?
\end{flushitemize}

In order to evaluate these questions, we conducted a between-subjects user study in which human participants took part in a mock manufacturing assembly activity. Participants were given construction tasks while a robot with tasks of its own would occasionally interrupt them to request assistance. The study had three conditions in which we varied the mechanism used by the robot to select an appropriate moment to interrupt the participant.

\paragraph*{Random interruptions (\textbf{RND})} The robot interrupted participants after it waited for a random amount of time, reflecting the current behavior of interruptibility-unaware robots. For example, the robots evaluated by Mutlu and Forlizzi~\cite{Mutlu2008} operated in the same environment as hospital staff, interrupting them randomly to gain attention as needs arose. In our study, the robot's algorithm tried to emulate this behavior by randomly selected a wait time from a uniform distribution in the range [0,30] sec; after which, it flipped a fair coin every 0.5 sec to decide whether to interrupt.  Wait times in the study ranged from 2 to 37 sec.

\paragraph*{Wizard-of-oz interruptions (\textbf{WOZ})} The robot interrupted participants when a human (wizard) signaled it was an appropriate time. Wizards were provided with a real-time video feed from the robot's camera and, during pilot trials, were instructed to make moment-by-moment decisions to interrupt the participant or not, simulating the decision made by our interruptibility models\savefootnote{foot:wizard-instructions}{The wizards were asked to (1) treat images from the video at each moment as a static image to decide whether they would interrupt the participant at that moment, (2) specifically ignore the screen on the participant's tablet and the task schedules that they were becoming accustomed to (to the extent possible), and (3) give the robot and human tasks equal importance.}. Once the decision to interrupt was made, the wizards could perform no more actions until the next robot incursion into the study space. During study trials, there was no interaction between the experimenters and the wizard. We recruited two wizards and observed that, despite similar instructions, differing social norms and attitudes among individuals led one wizard to be more conservative in their interruptions than the other. We therefore had each wizard participate in 50\% of WOZ trials to help account for this effect.

\subsubsection*{Model-Based interruptions (\textbf{MDL})} the robot interrupted participants based on output from an interruptibility classifier using a Latent-Dynamic Conditional Random Field~\cite{Morency2007} (LDCRF), as introduced in \cite{banerjee2017temporal}. We chose this method because (1) it directly addresses the problem of classifying interruptibility\savefootnote{foot:interruptibility-vs-engagement}{As we state in \cite{banerjee2017temporal}, ``... classifying interruptibility poses its own research problem because interruptibility can be high even when the person shows neither intent-to-engage nor awareness of the robot.''}, and (2) it uses features from the social cues of interruptibility projected by a human and the contextual cues of interruptibility obtained from objects in the scene.

\smallskip
In the following sections, we first describe our computational framework to enable online interruptibility classification using the LDCRF. We then present the design of and results from the user study to answer the above research questions.

\section{Computational Framework}
\label{sec:computational-framework}

\begin{figure}[t]
\centering
\includegraphics[width=\linewidth]{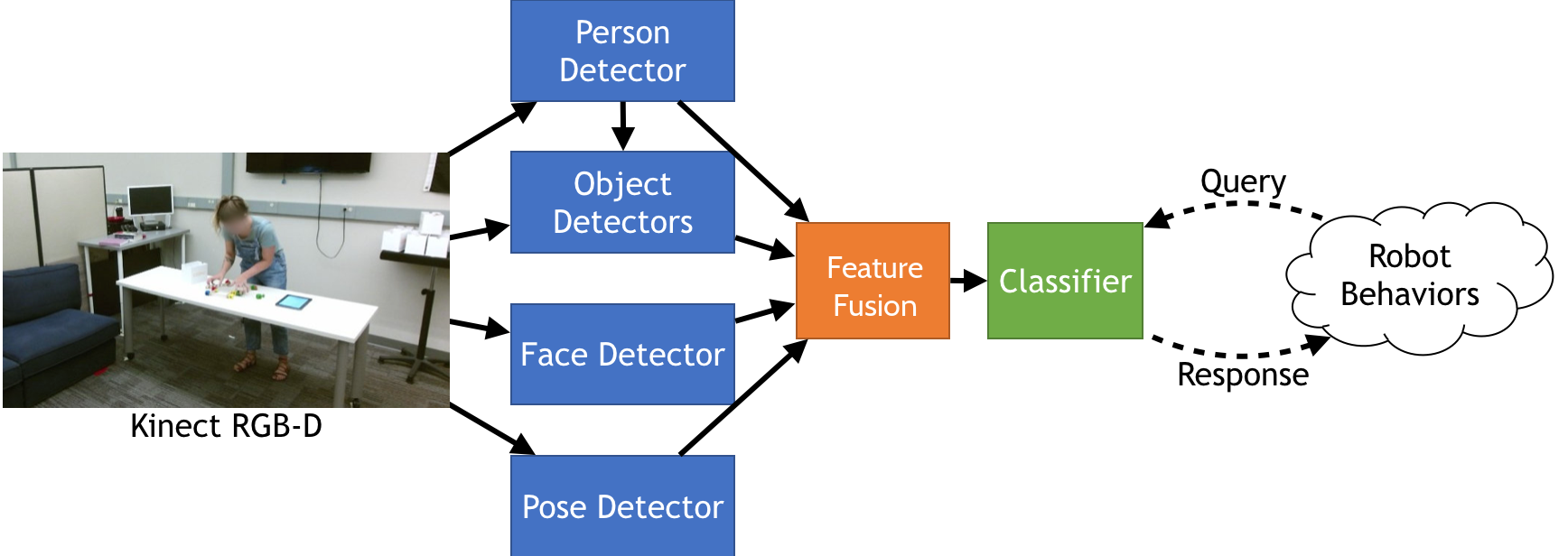}
\vspace{-.5cm}
\caption{Classification pipeline for interruptibility.}
\label{fig:classification-pipeline}
\vspace{-.6cm}
\end{figure}

Our goal is to enable a mobile robot to effectively classify a person's state as interruptible or not interruptible based only on data available from onboard sensors and without any information about the person's schedule or current job assignment.  We establish our approach based on our prior work~\cite{banerjee2017temporal}, in which we compared multiple temporal models on the task of classifying interruptibility using only LIDAR and camera data, obtaining the best performance with an LDCRF on a fixed image dataset.  However, applying the LDCRF from~\cite{banerjee2017temporal} in our pilot studies and comparing the output of the model to hand-annotated interruptibility labels resulted in relatively low performance with an F1 score of 0.79, compared to the near-perfect MCC scores in the original paper. More importantly, the model was very inconsistent in its predictions, frequently oscillating between labels on a relatively static scene because it was exposed to a much more diverse set of human poses and tasks in our study than in the controlled kitchen dataset used in the original work.  Additionally, \cite{banerjee2017temporal} relied on offline person recognition and hand-coded object labels, which were not practical for use on an actively deployed mobile robot.  Thus, below we present our online perception system, which uses features inspired by~\cite{banerjee2017temporal} with the addition of pose estimation features, resulting in an improved F1 score of 0.87 and very consistent classification.  We then outline our LDCRF training process.  Our full pipeline is visualized in Fig.~\ref{fig:classification-pipeline}.

\subsection{Perception System}
\label{sub:framework-perception}

The perception system of the robot (1) detects people in the scene, (2) uses a series of detectors to analyze the state of each individual, and (3) merges the output of the detectors into a feature vector for processing by the classification model.

\subsubsection*{\textbf{Person Detector}} We use the You Only Look Once (YOLOv2) ~\cite{redmon2016yolo9000} deep neural network to detect people in the scene. This detector was chosen for ease of use and setup, and for its accuracy and speed. It never missed a person in our user study, and published person detections at $>$10fps.

\subsubsection*{\textbf{Feature Detectors}} Once a person has been identified in the scene, we employ several deep networks to extract interruptibility-relevant features about the person. We include features from the prior work, such as the coarse gaze estimate of a person and the objects associated with them, and introduce the skeletal data for improved classification.

\underline{Face Detector}: We use a cascaded deep network~\cite{zhang2016joint} for face detection and coarse gaze estimation. The detector returns facial keypoints, which we translate into an enumerated gaze estimation variable. Features:  \textit{at\_robot}\textbar\textit{left\_right}\textbar\textit{down} (Enum). Framerate: 7-10fps.

\underline{Object Detector 1}: We use another implementation of YOLOv2 that runs over higher resolution images which are cropped to include regions around people in the scene---information that is obtained from our person detector. This detector was trained on MSCOCO~\cite{lin2014microsoft} and returns counts of objects and their positions.  Features: \textit{book}, \textit{bottle}, \textit{bowl}, \textit{cup}, \textit{laptop}, and \textit{cell phone}. Framerate: $>$10 fps.

\underline{Object Detector 2}:  We use Faster-RCNN~\cite{ren2015faster} fine-tuned to identify study-related objects on the table, in our case the tablet participants used throughout the study. As with our other object detector, this returns counts and positions of detected objects. Features: \textit{tablet}.  Framerate: $>$10 fps.

\underline{Pose Detector}: We use a convolutional pose machine (CPM)~\cite{cao2016realtime} to infer a person's skeletal keypoints. These keypoints are then refined into joint angles and vectors for our classifier. Features: \textit{nose\_vec\_x}\textbar\textit{y}, \textit{angle\_left}\textbar\textit{right}: \textit{elbow}, \textit{wrist}, \textit{shoulder}, \textit{eye}.  Framerate: 5-7 fps.

\subsubsection*{\textbf{Feature Fusion}} Each of the above detectors runs in parallel and at different rates.  The Feature Fusion module used Euclidean distance heuristics, as in the prior work, to aggregate the output of the various detectors into a single feature vector describing the most up-to-date estimate of the scene.

\begin{figure}[t]
\centering
\includegraphics[width=.45\linewidth]{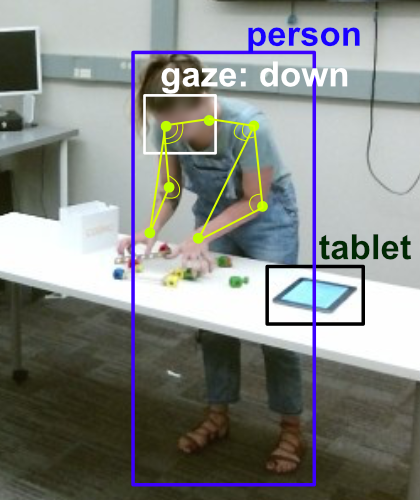}
\includegraphics[width=.45\linewidth]{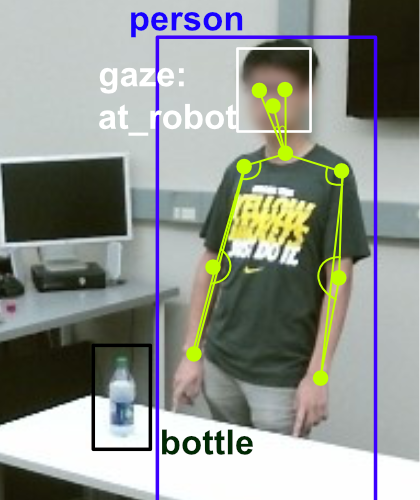}
\caption{Visualization of our detected and used features}
\label{fig:feature-visualization}
\vspace{-.6cm}
\end{figure}

\subsection{Classification Model}

The LDCRF is an undirected graphical model that uses local feature functions to associate a temporal sequence of feature vectors to a sequence of classifications. We adapt the implementation described in \cite{banerjee2017temporal} to fit our needs.

\subsubsection*{Implementation}
We used the feature functions defined in \cite{banerjee2017temporal}, which required the specification of two hyperparameters: the number of hidden states in the LDCRF ($|\mathcal{H}_{y_i}|$) and a time window of observation in the temporal sequence ($\omega$). We implemented the feature functions with the HCRF library\savefootnote{foot:hcrf}{\url{https://sourceforge.net/projects/hcrf/}} and the parameters for the model were trained with BFGS gradient descent. To improve classification, prior to training, testing, and inference on the robot, missing data in the feature vectors were imputed by propagating the last known valid values\savefootnote{foot:feature-propagation}{A buffer of 4 secs of feature vectors was used for imputation on the robot.}. Feature values were then normalized to a maximum absolute value of 1, and non-imputable $NaN$ values were ignored.

\subsubsection*{Model Training}
We trained the model on data collected over the course of 4 pilot runs and the RND condition. The feature vectors in the data were annotated by two of the coauthors of this paper on a binary scale of interruptibility, with 0 as uninterruptible and 1 as interruptible\savefootnote{foot:annotation-procedure}{The coders operated under the instructions provided to wizards in the WOZ condition: they observed a video feed from the robot's camera and provided a moment-by-moment label of whether the participant was interruptible.}. Cronbach's Alpha score of inter-rater reliability was 0.97. It was 0.95 for each coder compared to the ground truth of interruptibility (Sec.~\ref{sub:study-measurements}) of the participant obtained from the tablet. All hyperparameter configurations underwent five-fold cross validation, with special care undertaken to ensure that entire study trials were not shared between the train and test sets. We ultimately use $|\mathcal{H}_{y_i}| = 4$ hidden states and an $\omega = 3$ window of observation.

\section{Study Design}
\label{sec:study}

We conducted a between-subjects user study to evaluate the research questions outlined in Sec.~\ref{sec:overview}. The study involved 48 trial participants\savefootnote{foot:study-pilot-stats}{Six additional participants took part in pilot trials used to tune build complexity, robot behavior, and to familiarize the wizards with their interface.}.  Six trials were excluded from the study analysis: two due to hardware malfunction, and four due to participants deviating from the study protocol.  The resulting 42 participants (20 women, 22 men) were aged between 21 and 29  ($Mdn = 24$).  The study took approximately 50 min, and participants were paid \$10 USD.

\subsection{Study Procedure}
\label{sub:study-setup}

We devised a skill-based experimental task in which human participants took part in a mock manufacturing assembly activity.  Participants were instructed to construct structures (\textit{builds}) out of wooden pieces (Fig.~\ref{fig:setup_and_builds}b), and told that their build process would be video recorded to be used later as training data for the robot.  Additionally, participants were told that the robot was performing and studying its own builds, and that it would occasionally enter the space to request assistance.

\subsubsection*{\textbf{Pre-Study}}
Upon arrival, participants were briefed on the study, completed consent forms, and filled in a pre-study questionnaire.  Nearby, to support the narrative of the robot learning to construct builds, an experimenter could be seen ``training''\footnote{No actual training of the robot occurred during the study trials.} the robot by responding to the robot's questions (e.g., ``Is this a correct build?'').

\subsubsection*{\textbf{Study Space}}
\begin{figure}[t]
\centering
\includegraphics[width=\linewidth]{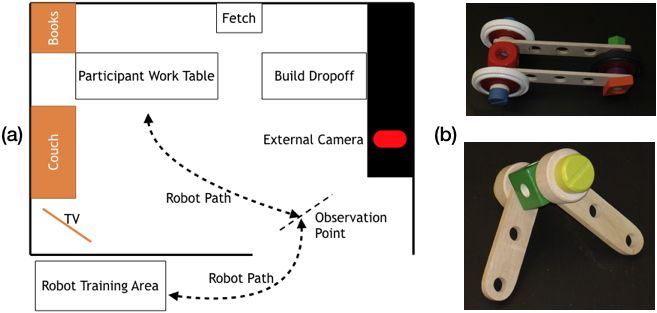}
\caption{(a) Map of study area, (b), Sample builds for participants: (Top) Main Build, (Bottom) Interruption Build}
\label{fig:setup_and_builds}
\vspace{-0.4cm}
\end{figure}

After the study briefing, participants entered the building area (Fig.~\ref{fig:setup_and_builds}a), consisting of an enclosed space with fetch area for retrieving build components, a work area for construction, and a dropoff area for completed builds.  A key element of the study design is that the study schedule was split into periods of work and leisure to ensure that participants had periods of low and high interruptibility.  To induce participants to showcase a diverse range of natural leisure behaviours (to fully evaluate the performance of the classifier and generalizability of our system), the room included a TV playing muted videos\savefootnote{foot:wired-playlist}{\url{http://bit.ly/2xR65aG}}, a stack of books, and a couch.  Participants were also allowed to keep their cell phones.  Overall, during breaks 64\% sat on the couch, 50\% used their cell phones, 40\% drank a refreshment, and 14\% read a book.

\begin{figure}[t]
\centering
\includegraphics[width=\linewidth]{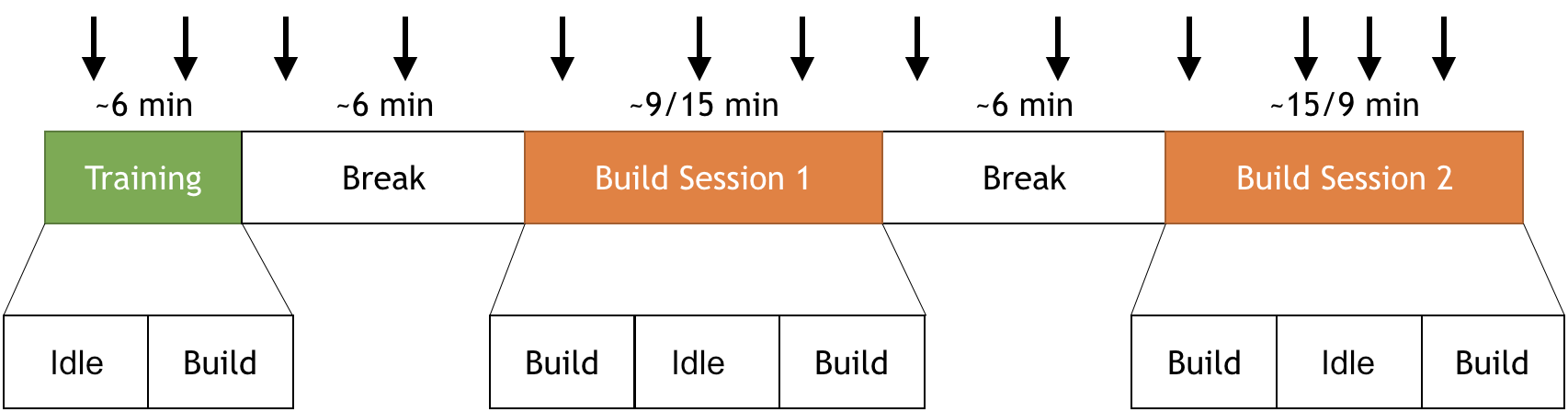}
\caption{Sample timeline of a trial, arrows indicate interruptions}
\label{fig:trial-timeline}
\vspace{-0.6cm}
\end{figure}

\smallskip
For the remainder of the study period, participants alternated between constructing builds (\textit{build}) and break times (\textit{idle}), while being occasionally interrupted by the robot.  Fig.~\ref{fig:trial-timeline} presents an example timeline.

\smallskip
\subsubsection*{\textbf{Builds}}
Each participant trial consisted of 3 build sessions.  The first build session was a training session during which participants were allowed to ask questions and acclimate themselves to the task and the robot. We do not report data from this session.  Sessions 2 and 3 each consisted of two builds, with a short break in between.  Instructions for each build were provided on a tablet located on the work table; the tablet remained blank until the designated build time.  The build sessions were either 15 min or 9 min in length, and were presented to all participants in a counterbalanced manner. The different length build sessions were configured to provide differing degrees of time pressure on the participant. In addition, pilot studies showed that some participants improved in build performance due to learning; the counterbalanced sessions were used to amortize the effects of learning on high time pressure and low time pressure sessions. All builds in a build session had a time limit, and participants were shown a countdown timer 30 sec before the end of this time limit; participants were not allowed to work past the end of the limit.

\subsubsection*{\textbf{Breaks}}
Each trial included two break times approximately 6 min in length (differences in duration occurring due to robot interruptions), during which the tablet was taken away and the participants were invited to rest on the couch.  The purpose of the break was to expose the robot to interruptible human behavior.  In both cases, the experimenters presented fictitious excuses to the participant for pausing the study, in one case claiming a non-existent tracking device required adjustment, and in the other case simulating a tablet malfunction.  For both breaks, experimenters explained the pause in the experiment, invited participants to wait on the couch, and then returned at the end of the break to ``continue'' the study.   Participants were told that the robot interruptions would continue since the robot remained unaffected by the glitch.

\subsubsection*{\textbf{Robot Interruptions}}
The robot continually entered the building area looking for assistance from the start to the end of a trial. The schedule of these entrances was not predefined and the robot was sent back in as soon as it returned from an interruption. The first three robot entrances coincided with the training build session and part of the first break; we allowed participants to ask questions during these interruptions and do not report data from them. The robot was equipped with a small box containing the blocks for its builds and a tablet, which provided instructions to the robot builds.

During an entrance, the robot followed the path shown in Fig.~\ref{fig:setup_and_builds}a. It waited at the observation point upon entering and after waiting---a random duration in RND, until 2.5 sec of consecutive\savefootnote{foot:int-request-rate}{Empirically, 2.5 sec of consecutive classifications at 2 Hz worked well.} interruptible classifications in MDL, or until the wizard sent an interruptible signal in WOZ---chose to move toward the participant. Upon arrival, the robot verbally requested assistance and waited for 2 min. Participants were aware of the wait duration and could accept the interruption within the time limit by grabbing the tablet, at which point the robot waited indefinitely until the build was completed. If the participants did not respond in 2 min, the robot left the participant build area. Upon returning to the training area, the robot audibly requested verification of the build (e.g., ``Is this a correct build?'') from an experimenter. The experimenter provided a Yes/No response on whether the interruption was built, prepared the next robot build, and sent the robot back.

\subsubsection*{\textbf{Post-Study}}
After the last build session, participants were asked to complete a post-study questionnaire, were debriefed on the purpose of the study and the deceptions that we used.

\subsection{Hypotheses}
\label{sub:study-hypotheses}

Our central premise is that the robot in MDL \& WOZ will interrupt at appropriate moments, and that such interruptions will improve robot task performance and the social perceptions of the robot compared to those metrics in RND. Based on results from HFE research~\cite{Kolbeinsson2017}, we predict that there will be no effect on human task performance. Specifically, we formulate:

\begin{flushitemize}
\item[\textbf{H1} \small{(RQ1)}] With an interruptibility classifier (MDL), the robot will interrupt fewer builds than it would without the classifier (RND), waiting longer to interrupt when participants are building and interrupting more quickly when they are idle. In addition, the robot with the classifier will interrupt as many builds as a robot directed by a human (WOZ).

\item[\textbf{H2} \small{(RQ2)}] When the robot has interruptibility-aware behavior (MDL \& WOZ), participant task performance will be no better than participant task performance when a robot interrupts at random (RND).

\item[\textbf{H3} \small{(RQ3)}] When the robot has interruptibility-aware behavior (MDL \& WOZ), fewer of its tasks will be ignored and it will not need to spend as much time awaiting human assistance as it will when it interrupts at random (RND).

\item[\textbf{H4} \small{(RQ4)}] Participants will perceive an interruptibility-aware robot (MDL \& WOZ) as more socially aware and considerate than one that interrupts at random (RND).
\end{flushitemize}

\subsection{Measurements}
\label{sub:study-measurements}


Prior work in HCI and HFE quantifies task performance using metrics such as time on task~\cite{McFarlane2002,Adamczyk2004,Mark2008,Kolbeinsson2017}, the number of tasks completed~\cite{McFarlane2002}, and task switching time~\cite{Iqbal2006,Kolbeinsson2017}. We use similar quantitative measures of human and robot performance, and 5-point Likert scale responses to questions of participant opinions and participant background:

\begin{figure*}[t]
\centering
\includegraphics[width=0.69\linewidth]{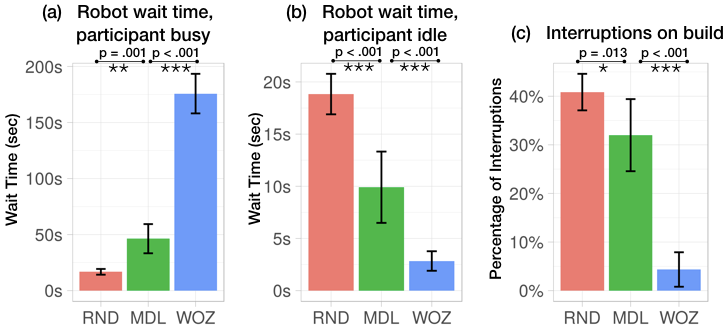}
\caption{Data and analysis for results in Sec.~\ref{sub:results-rq1}. In Fig.~\ref{fig:model-works},~\ref{fig:human-task-completion},~\ref{fig:robot-task-completion}~\&~\ref{fig:social-aptitude}, asterisks indicate level of statistical significance after post-hoc tests: * $p < 0.05$, ** $p < 0.01$, ***$ p < 0.001$. Error bars in the bar charts indicate the 95\% confidence interval.}
\label{fig:model-works}
\end{figure*}

\begin{flushitemize}
\item[\textbf{M1} \small{(RQ1)}] Percentage of builds interrupted by robot; robot wait (to interrupt) time when participant is on build; robot wait (to interrupt) time when participant is idle.
\item[\textbf{M2} \small{(RQ2)}] Participant's time idle; total number of tasks done.
\item[\textbf{M3} \small{(RQ2/RQ3)}] Number of interruptions of the participant; number of interruptions ignored; interruption lag, measured as the time between when the robot requests assistance and the participant begins constructing the robot build; interruption duration, measured as the total time the robot waits after it has requested assistance.
\item[\textbf{M4} \small{(RQ4)}] Perceived appropriateness of timing\savefootnote{foot:good-time-question}{Q: When the robot interrupted you, was it a good time to interrupt?}; perception of robots considerateness (workload-awareness)\savefootnote{foot:considerate-question}{Q: Did the robot take your workload into consideration when asking for help?}.
\item[\textbf{M5} \small{(Control)}] Experience with building blocks; proficiency at multitasking; familiarity with robots; motivation and anxiety during trial; difficulty of trial; predictability of robot interruptions.  These measures instrumented factors that had the possibility to confound results based on results in prior literature and our experience from the pilot studies.
\end{flushitemize}

Most quantitative measures were automatically logged from timestamps on the tablet and the robot, but some discrepancies caused by unexpected participant behavior\savefootnote{foot:unexpected-behavior}{For example, ignoring a build on the main tablet, or picking up the robot tablet and then replacing it without completing the robot build} were corrected using video from the external camera. For all trials, timestamps from the tablets are treated as ground truth of participant interruptibility. In addition to the above metrics, we also allowed participants to verbally elaborate on their choices and reasoning during post-study debriefing.

\section{Analysis of Model-Driven Robot Behavior}
\label{sec:classifier-analysis}

In this section, we evaluate the performance of the robot's interruptibility model in the study setting and explore metrics pertaining to the question, ``Can an integrated system be developed to accurately estimate human interruptibility online on a robot platform?'' (\textbf{RQ1}). Our analyses in this section are conducted using a one-way analysis of variance (ANOVA) with the study condition as an independent variable. The ANOVA is followed by post-hoc comparisons using Tukey's HSD. Results for this section can be seen in Fig. ~\ref{fig:model-works}.

\subsection{Results}
\label{sub:results-rq1}

We first examine the amount of time the robot waited at the observation point when participants were busy or idle as an indication of moment-to-moment interruptibility classifier accuracy. Concretely, we expect an accurate classifier to make the robot wait longer when a participant is busy, and not as long when the participant is idle. Over the course of the 42 trials, the robot entered the manufacturing environment 389 times (161 during build times, and 228 during break times). Of the 161 observations when participants were busy, the data shows a significant difference between the conditions ($F(2,158)=189, p=1.1e^{-42}$), with the robot waiting longer in MDL ($M=46.3, SD=48.2$) than in RND ($M=16.8, SD=9.73$), and longer in WOZ ($M=176, SD=60.8$) than in MDL (Fig.~\ref{fig:model-works}a). Of the 228 observations when participants were idle, there was again a significant difference between the conditions ($F(2,225)=38.7, p=3.5e^{-15}$), with the robot waiting less time in MDL ($M=9.91, SD=15.6$) than in RND ($M=18.8, SD=8.78$), and less time in WOZ ($M=2.84, SD=3.75$) than in MDL (Fig. ~\ref{fig:model-works}b)

\begin{figure*}[t]
\centering
\includegraphics[width=0.69\linewidth]{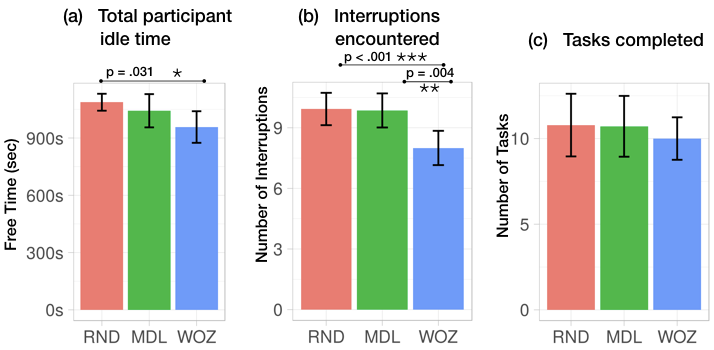}
\caption{Data and analysis for results in Sec.~\ref{sub:results-rq2}.}
\label{fig:human-task-completion}
\end{figure*}

We next examine the percentage of interruptions per trial that occurred during a build. We expect that a more accurate interruptibility classifier will have a lower percentage of interruptions in the middle of a build. As shown in Fig.~\ref{fig:model-works}c, the data from the 14 trials indicates a significant difference between the conditions ($F(2,39)=74.8, p=4.5e^{-14}$), where the percentage is lower in WOZ ($M=0.043, SD=0.061$) than in MDL ($M=0.32, SD=0.13$), and lower in MDL than in RND ($M=0.41, SD=0.065$).

We observe significant differences between our two wizards in the above metrics. The conservative wizard (wizard \textbf{C}) never interrupted a participant in the middle of a build ($M=0, SD=0$), while the aggressive wizard (wizard \textbf{A}) preferred to interrupt as a participant completed their task, sometimes catching them at the end of a build ($M=0.087, SD=0.061$). As a result, the robot's wait time at the observation point differs between the wizards. However, both wizards' metrics are closer to each other than to MDL or RND.

\subsection{Summary and Discussion}
\label{sub:discuss-rq1}

Our results support \textbf{H1}. In this study, we defined an appropriate interruption as one that occurs when the participant is idle, and not engaged on a tablet build.  The above analyses show that the classification model results in appropriately timed interruptions, with a model-equipped robot approaching participants quickly when they are free while waiting to approach when they are busy. Examining these metrics, it is clear that a robot equipped with an interruptibility model is socially aware through its ability to use the model to autonomously select appropriate times to engage with people. However, the robot controlled by a wizard is the most interruptibility-aware, indicating that we still have room to improve the model in order to achieve human-level accuracy.

\section{Analysis of Human Task Performance}
\label{sec:human-performance-analysis}

The results above validate that an interruptibility-aware robot has an increased likelihood of making appropriately timed interruptions. In this section, we explore the effects that this change in robot behavior has on human task performance.  Specifically, we examine the metrics relevant to, ``How does interruptibility-aware robot behavior affect human task performance when a robot regularly needs assistance?'' (\textbf{RQ2}). To control for Type II errors in establishing the \textit{no difference} hypothesized by \textbf{H2}, we follow the guidelines suggested by \cite{julnes1989analysis}, in particular an alpha level of 0.25 (i.e., $p>.25$) to claim \textit{no difference} and use the term \textit{indeterminate} for alpha levels between 0.05 and 0.25 (i.e., $.25 < p < .05$).

\subsection{Results}
\label{sub:results-rq2}

We find that the self-reported rating of experience with building blocks (\textit{build experience}) was a significant confounding factor in participant build proficiency. Correlating self-reported experience to observed task performance, we observe most difference between those who self-reported experience as 1 or 2 (\textit{low} experience), and those who reported experience of 3 or higher (\textit{high} experience).  Participants with \textit{high} and \textit{low} experience were similarly distributed between conditions, with 10 \textit{high}, 4 \textit{low} experience participants in RND and MDL, and 9 \textit{high}, 5 \textit{low} experience participants in WOZ. The following analyses control for build experience.

\subsubsection*{Idle Time}
We assume that moments when the participant is idle are moments of lost productivity; even while the main builds are unavailable, the robot has tasks that can be completed. We therefore wish to minimize participant idle time. For the 14 trials in each condition, a two-way ANOVA with the study condition and build experience as independent variables shows a significant effect of study condition ($F(2,36)=3.36, p=.046$) and an indeterminate effect of build experience ($F(1,36)=1.95, p=.17$). A post-hoc Tukey's HSD reveals lower idle time in WOZ ($M=957, SD=143$) than in RND ($M=1087, SD=76.7$), but no difference between MDL ($M=1043, SD=151$) and RND and an indeterminate difference between MDL and WOZ (Fig.~\ref{fig:human-task-completion}a).

\subsubsection*{Interruptions Encountered}
In our study, the robot continually re-entered the building area to interrupt, which resulted in participants that attended to interruptions quickly receiving more interruptions. Therefore, the number of interruptions presented to a participant is an indication of the number of tasks that they encountered\savefootnote{foot:total-tasks-encountered}{All participants received 4 tasks from the main tablet.}: another indicator of task performance. For the 14 trials in each condition, a two-way ANOVA with study condition and build experience shows a significant effect of study condition ($F(2,36)=7.63, p=.0017$) and an indeterminate effect of build experience ($F(1,36)=2.04, p=.16$). A post-hoc Tukey's HSD reveals a lower number of interruptions encountered in WOZ ($M=8, SD=1.47$) than in RND ($M=9.93, SD=1.38$) and a lower number in WOZ than in MDL ($M=9.86, SD=1.46$), with no significant difference between MDL and RND (Fig.~\ref{fig:human-task-completion}b).

\subsubsection*{Interruptions Ignored}
\begin{figure*}[t]
\centering
\includegraphics[width=0.69\linewidth]{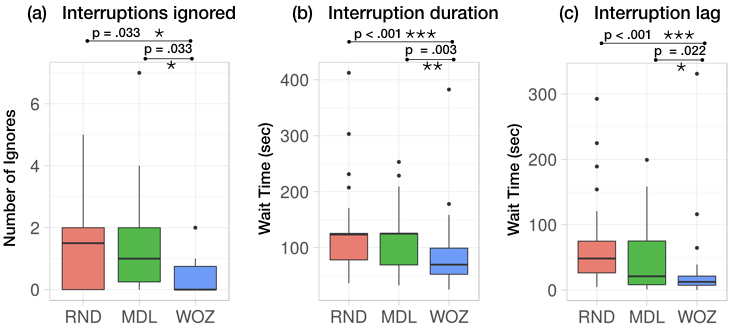}
\caption{Data and analysis for results in Sec.~\ref{sub:results-rq3}.}
\label{fig:robot-task-completion}
\end{figure*}

Participants were given the freedom to ignore robot interruptions during the study. We expected such ignores to occur when participants were overwhelmed, and therefore consider the number of interruptions ignored as a negative indicator of human task performance. For the 14 trials in each condition, a Kruskal-Wallis test shows a significance of study condition ($H(2)=7.15, p=.028$) and an indeterminate effect of build experience ($H(1)=2.95, p=.086$). A post-hoc pairwise Wilcoxon rank sum test with Benjamini \& Hochberg~\cite{benjamini1995controlling} correction reveals a lower number of interruptions ignored in WOZ ($Mdn=0$) than in RND ($Mdn=1.5$) or MDL ($Mdn=1$), with no difference between MDL and RND (Fig.~\ref{fig:robot-task-completion}a).

\subsubsection*{Tasks Completed}
The final measure is the total number of tasks (builds + robot interruptions) that were completed by participants during a trial. For the 14 trials in each condition, a two-way ANOVA with study condition and build experience as independent variables finds a significant effect of build experience ($F(1,36)=14.9, p=.0004$) and no difference with study condition ($F(2,36)=0.22, p=.8$) (Fig.~\ref{fig:human-task-completion}c).

\medskip

We again comment on the differences between our wizards. Although there is no difference between the wizards in the amount of idle time, interruptions encountered, and tasks completed, the wizard \textbf{C}'s interruptions were ignored less often ($Mdn=0$) than wizard \textbf{A}'s ($Mdn=1$).

\subsection{Summary and Discussion}
\label{sub:discuss-rq2}

The results above lead us to mixed conclusions regarding \textbf{H2}. Firstly, we find that idle time is minimized by the awareness of interruptibility, with participants exposed to the perfect interruptibility-aware robot in WOZ enjoying significantly less idle time than participants in RND. An obvious cause of the reduced idle time is the robot's behaviour in waiting to interrupt participants until they are free (Sec.~\ref{sec:classifier-analysis}), which in turn causes participants in WOZ to encounter fewer tasks than participants in either MDL or RND. However, we find that waiting until participants are free leads to fewer interruption builds that are ignored, thereby offsetting the potential cost to throughput incurred by presenting fewer tasks to people. Ultimately, we find the factors relating to task throughput balance each other such that the total number of tasks completed by humans is not significantly different due to interruptibility-aware behaviour.

The tradeoff in the factors affecting task completion explain the similar throughput between RND and WOZ, but they fail to explain the similarity in the task metrics between RND and MDL, despite the results in Sec.~\ref{sec:classifier-analysis} showing that the robot tended to wait longer and interrupted fewer builds in MDL. This discrepancy is explained instead by the results from HFE research \cite{Lee2015, Kolbeinsson2017}, which suggest the embodiment of the robot interruptions and the skill-based main task contributed to unaffected task performance. Therefore, it is likely that despite more mistimed interruptions in RND than in MDL, participants were able to optimize their build process such that their performance remained unaffected on the metrics of task throughput that we instrumented.

In conclusion, we find that all three of our conditions achieved similar task throughput, suggesting our participants maximized their potential throughput in our manufacturing environment. However, the maximization came at the cost of robot tasks being ignored in the interruptibility-unaware condition of RND. In fact, we find that the addition of interruptibility-aware behaviour (WOZ in particular) greatly improved the efficiency of the robot, particularly with a reduction in the number of its tasks that were ignored. This is explored further in the next section.

\section{Analysis of Robot Task Performance}
\label{sec:robot-performance-analysis}

In this section, we examine metrics relevant to answering the question, ``How does interruptibility-aware robot behavior affect robot task performance when relying on humans for assistance?'' (\textbf{RQ3}). In evaluating our hypothesis (\textbf{H3}), we make a distinction between the time spent by the robot waiting at the observe point, and the time spent by the robot waiting in front of the participant's work table. We do not consider the observe time to be wasted time, as we assume that the robot might find an alternative interruption candidate during this time in a different environment.

\subsection{Results}
\label{sub:results-rq3}

Our conclusions are drawn from the number of interruptions that the robot presented (Fig.~\ref{fig:human-task-completion}b), the number of those that were ignored (Fig.~\ref{fig:robot-task-completion}a), and the delays incurred by the robot by waiting on the human after it requested assistance.  For the last metrics, we only present analyses on interruptions initiated during a build\savefootnote{foot:wait-on-observe-meaningless}{The difference between the study conditions is most apparent in such interruptions. Kruskal-Wallis tests on robot delay data from the interruptions when participants were observed idle show no significant effect of the study condition and show instead a significant effect of participant build experience.}. Our analyses use a Kruskal-Wallis test on study condition followed by post-hoc pairwise Wilcoxon rank-sum tests with Benjamini \& Hochberg correction.

The interruption duration is unproductive robot time spent waiting on the human's assistance, and is therefore a measure of low productivity. We hypothesize that poorly timed interruptions result in a longer interruption duration, and therefore more time wasted by a robot that needs assistance. Of the 161 interruptions examined, the data reveals a significant difference between the study conditions ($H(2)=15, p=.0006$), with a shorter interruption duration in WOZ ($Mdn=69.6$) than in MDL ($Mdn=125$) or in RND ($Mdn=123$) (Fig.~\ref{fig:robot-task-completion}b).

The interruption lag is another metric of how long the robot had to wait on participants, and is a better indicator of the effect of appropriate timing to the robot's task delay because it is not affected by a participant's capability to build, or by whether the interruption was ignored. As with interruption duration, higher interruption lag means more time wasted by a robot and lower efficiency. Of 127 interruptions, the data reveals a significant effect of study conditions ($H(2)=24.8, p=4.2e^{-6}$), with lower lag in WOZ ($Mdn=12.4$) than in MDL ($Mdn=21.0$) or in RND ($Mdn=48.2$) (Fig.~\ref{fig:robot-task-completion}c). We also observe a marginal ($p=0.08$) reduction in lag between MDL and RND.

We observe a significant difference between our wizards in both of the above metrics, with participants showing lower interruption lag and lower interruption duration with wizard \textbf{C}. Although our wizards are closer to each other than to MDL or RND on the duration, the interruption lag of participants to wizard \textbf{A} ($Mdn=20.7$) is closer to MDL, than to wizard \textbf{C} ($Mdn=9.6$).

\subsection{Summary and Discussion}
\label{sub:discuss-rq4}

Our results support \textbf{H3}, and indicate that interruptibility awareness has a positive impact on robot task performance. Not only is the robot able to accomplish the same amount of work with fewer requests for assistance, but well-timed interruptions also reduce the amount of time the robot has to wait on the participant to respond to its request, even when the interruptibility-awareness might not be perfect (as in MDL). In summary, well-timed interruptions allow a robot to operate more efficiently, completing tasks with fewer requests and in less time. In the next section, we evaluate participants' perception of such well-timed interruptions.

\section{Analysis of Robot Impressions}
\label{sec:perception-analysis}

\begin{figure}
\centering
\includegraphics[width=0.6\linewidth]{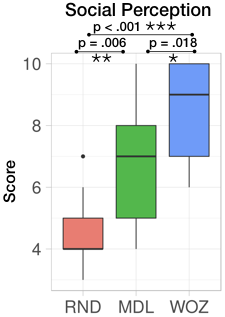}
\caption{Data and analysis for results in Sec.~\ref{sub:results-rq4}.}
\label{fig:social-aptitude}
\vspace{-.5cm}
\end{figure}

Our results thus far show that the robot in the MDL and WOZ conditions succeeded in interrupting participants more appropriately, that this behavior did not have significant impact on participant task performance, but that it did improve robot task performance.  Here, we evaluate \textbf{H4} using the Likert scale responses to the two questions, \textbf{M4} (Cronbach's $\alpha = 0.7$), to comment on the impact of appropriate interruption timing to participants' opinions. For our analyses, we use a Kruskal-Wallis test on study condition followed by post-hoc pairwise Wilcoxon rank-sum tests with Benjamini \& Hochberg correction. We also drop one of the 14 responses in the MDL condition because the participant spent less than 10 sec on the post-study questionnaire.

\subsection{Results}
\label{sub:results-rq4}

For the 14 trials in each condition, the data reveals a significant difference ($H(2)=21.1, p=2.6e^{-5}$) in the scores of social perception between all three conditions, with participants rating WOZ ($Mdn=9$) the highest, followed by MDL ($Mdn=7$), followed by RND ($Mdn=4$). We did not observe any difference between our two wizards on the score (Fig.~\ref{fig:social-aptitude}).


\subsection{Summary and Discussion}
\label{sub:discuss-rq4}

Our results support \textbf{H4}; participants rated WOZ interruptions as better timed than MDL or RND, and found the robot in MDL to be more considerate than in RND.

\section{Insights and Conclusions}
\label{sec:conclusion}

In this paper, we have described the first fielded mobile robotic system capable of classifying human interruptibility based on social and contextual cues and without reliance on external sensors.  Our results supporting \textbf{H1} show that the system is effective at predicting interruptibility at high accuracy, and that using it our robot interrupts at more appropriate times than a robot without interruptibility awareness.  Results from our user study validate that developing interruptibility-aware robotic systems is important to future deployments of interactive autonomous systems.  Similar to prior work in HFE, we find that human performance of skill-based tasks is not affected by interruptions (\textbf{H2}), primarily because of participants effectively regulating their workload by ignoring the robot when too many tasks are given.  Critically, however, interruptibility-aware behavior improves metrics associated with robot task performance (\textbf{H3}) by reducing the robot's time wasted on inappropriate interruptions, and improves social perception of the robot (\textbf{H4}).

Additionally, our research highlights some of the complexities associated with interruptibility, such as the fact that even the two wizards in our WOZ condition, who underwent identical training and instruction, did not entirely agree on the appropriate timing of interruptions.  Thus, many factors beyond just social and contextual cues play a role in interruption timing, such as differences in personality, or simply the urgency of the task needing attention, and these should be explored in future research.  Continuing work is also needed to explore the difference in interrupting skill-based vs cognitive tasks, and to model the optimal way for a robot to behave during an interruption.

\section*{Acknowledgments}
This work was supported by an Early Career Faculty grant from NASA's Space Technology Research Grants Program.

\bibliographystyle{plainnat}
\bibliography{library_arx}

\end{document}